# Deep Learning for Computational Chemistry

Garrett B. Goh,[*,†] Nathan O. Hodas,[‡] Abhinav Vishnu[†]

[†]*High Performance Computing Group, Pacific Northwest National Laboratory, 902 Battelle Blvd, Richland, WA 99354*

[‡]*Data Science and Analytics, Pacific Northwest National Laboratory, 902 Battelle Blvd, Richland, WA 99354*

\* Corresponding Author: Garrett B. Goh

Email: garrett.goh@pnnl.gov

**Abstract**

The rise and fall of artificial neural networks is well documented in the scientific literature of both computer science and computational chemistry. Yet almost two decades later, we are now seeing a resurgence of interest in deep learning, a machine learning algorithm based on multilayer neural networks. Within the last few years, we have seen the transformative impact of deep learning in many domains, particularly in speech recognition and computer vision, to the extent that the majority of expert practitioners in those field are now regularly eschewing prior established models in favor of deep learning models. In this review, we provide an introductory overview into the theory of deep neural networks and their unique properties that distinguish them from traditional machine learning algorithms used in cheminformatics. By providing an overview of the variety of emerging applications of deep neural networks, we highlight its ubiquity and broad applicability to a wide range of challenges in the field, including QSAR, virtual screening, protein structure prediction, quantum chemistry, materials design and property prediction. In reviewing the performance of deep neural networks, we observed a consistent outperformance against non-neural networks state-of-the-art models across disparate research topics, and deep neural network based models often exceeded the "glass ceiling" expectations of their respective tasks. Coupled with the maturity of GPU-accelerated computing for training deep neural networks and the exponential growth of chemical data on which to train these networks on, we anticipate that deep learning algorithms will be a valuable tool for computational chemistry.

1.  **Introduction**

Deep Learning is the key algorithm used in the development of AlphaGo, a Go-playing program developed by Google that defeated the top human player in 2016.[1] The development of computer programs to defeat human players in board games is not new; IBM's chess-playing computer, Deep Blue, defeated the top chess player two decades ago in 1996.[2] Nevertheless, it is worth noting that Go is arguably one of the world's most complex board game. Played on a 19x19 board, there are approximately $10^{170}$ legal positions that can be played. Compared to the complexity of Go, it has been estimated that the Lipinski virtual chemical space might contain only $10^{60}$ compounds.[3,4]

Deep learning is a machine learning algorithm, not unlike those already in use in various applications in computational chemistry, from computer-aided drug design to materials property prediction.[5-8] Amongst some of its more high profile achievements include the Merck activity prediction challenge in 2012, where a deep neural network not only won the competition and outperformed Merck's internal baseline model, but did so without having a single chemist or biologist in their team. In a repeated success by a different research team, deep learning models achieved top positions in the Tox21 toxicity prediction challenge issued by NIH in 2014.[9] The unusually stellar performance of deep learning models in both predicting activity and toxicity in these recent examples, originate from the unique characteristics that distinguishes deep learning from traditional machine learning algorithms.

For those unfamiliar with the intricacies of machine learning algorithms, we will highlight some of the key differences between traditional (shallow) machine learning and deep learning. The simplest example of a machine learning algorithm would be the ubiquitous least-squares linear regression. In linear regression, the underlying nature of the model is known (linear in this context), and the input, otherwise known as the features of the model are linearly independent to each other. Additional complexity may be added to linear regression by transforming out the original data (i.e. squaring, taking the logarithm, etc.). As more of these non-linear terms are added, the expressive power of the regression model increases. This description highlights three characteristics of traditional (shallow) machine learning. First, the features are provided by a domain expert. In a process known as feature extraction and/or engineering, various transformations and approximations are applied, which can be motivated from first principles, or may be well-known approximations, or even educated guesses. Second, shallow learning is template matching. It does not learn a representation of the problem, it merely learns how to precisely balance a set of input features to produce an output. Third, its expressive power grows with the number of terms (i.e. parameters to be fitted), but it may require exponentially many terms if the nonlinear transformations are chosen poorly. For example, a simple power series expansion will require an extremely large amount of terms (and parameters) to fit functions with large amounts of oscillations.

The mapping of features into an output using the function provided is the task of a processing unit, and deep learning algorithms are constructed from a collection of these processing units, arranged in a layered and hierarchical fashion. Therefore, unlike simpler machine learning algorithms, it maps features through a series of non-linear functions that are stitched together in a combinatorial fashion to optimally maximize the accuracy of the model. As a result of this complex architecture, a deep learning algorithm learns multiple levels of representations that correspond to different levels of abstraction, forming a hierarchy of concepts. By constructing these so-called "hierarchical representations," deep learning has an internal state that may be transferred to new problems, partially overcoming the template matching problem. Finally, information can take many paths through the network, and, as a result, expressive power grows exponentially with depth.[10] Ultimately, these unique characteristics of deep learning enables it to utilize raw data directly as opposed to engineered features, and often the resulting models constructed produce a comparable level of predictive accuracy. Deep learning achieves this ability because within the multiple layers of the non-linear functions, the algorithm transforms the raw data and maps it to intermediate "output" that serve as input (features) for the latter layers in the algorithm, in the process gradually transforming raw data into learned features. In short, deep learning algorithms are potentially capable of automatically (i.e. without expert intervention) engineering the necessary features that are relevant to optimally predict the output of interest.

The majority of deep learning algorithms currently developed are based off artificial neural networks, and for the purpose of this review we will focus on deep neural networks exclusively. In the first half of this review, we will provide a brief non-technical introduction to deep learning, starting with a basic background on artificial neural networks and highlighting the key technical developments in the last decade that enabled deep neural networks. In addition, we will focus on how deep learning differs from traditional machine learning algorithms that are used in computational chemistry, and how the ongoing resurgence of deep learning differs from artificial neural network models in the 1980s, which may be regarded as its "parent" algorithm. In the next half of the review, we will include a survey of recent developments of deep learning applications across the field of computational chemistry, where we will examine its performance against existing machine learning models, and future prospects for contributing to the field. This review was written primarily to serve as an introductory entry point for computational chemists who wish to explore or integrate deep learning models in their research from an applications standpoint, and additional references to existing literature reviews will be provided to cover the more technical aspects of deep learning neural network architecture and optimization.

## 2. Deep Learning 101

Artificial neural networks (ANNs), on which most deep learning algorithms are based on, are a class of machine learning algorithm inspired by biological neural networks, used to estimate or approximate functions by translating a large number of inputs into a target output (Figure 1a). ANNs are constructed from a series of layers, and each layer comprises many "neurons". Each neuron accepts an input value from the previous layer, and maps it onto a non-linear function. The output of this function is used as the input for the next layer in the ANN, until it reaches the last layer, where the output corresponds to the objective that is to be predicted. In addition, a tunable parameter, the "weight" (or coefficient) of each neuron's function is adjusted in the construction of this model to minimize the error of the predicted value, a process known as "training" the neural network. Figuratively, the collection of these neurons in ANNs mimics the way neurons work in biological systems, hence its name, *artificial* neural network.

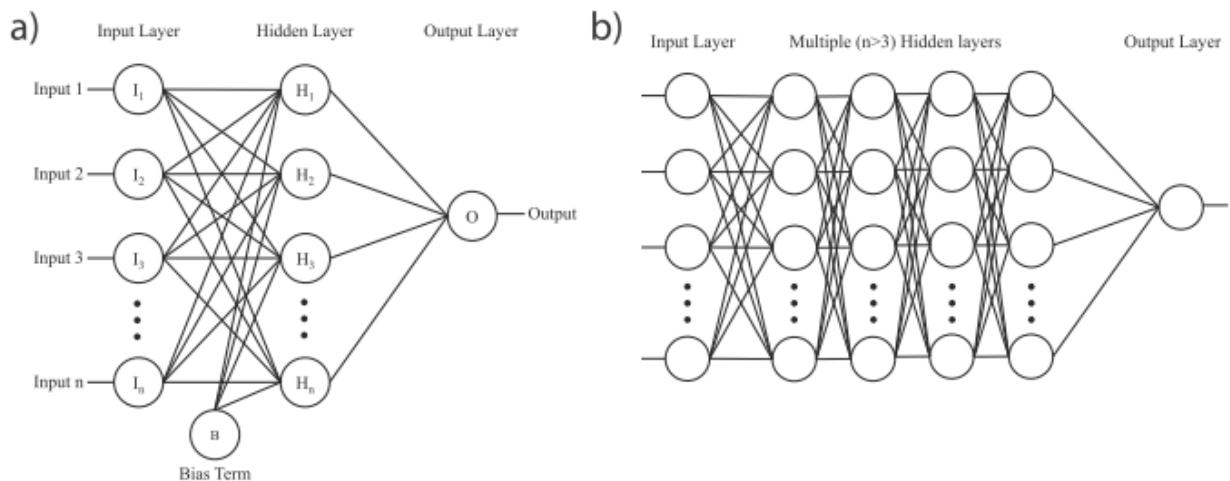

Figure 1: (a) Schematic representation of a traditional feedforward artificial neural network (ANN) with one hidden layer. Each neuron denoted as circles accepts a series of *n* input values and maps

it to an output using a non-linear function, with a bias term (i.e. output of the neural network when it has zero input) applied to all neurons in the hidden layer. (b) Deep neural network (DNN) differ from ANN by having multiple (n>3) hidden layers as depicted in the schematic diagram, the bias term is omitted here for simplicity.

The power of ANNs, as hinted earlier, lies in their ability to make multiple non-linear transformations through many hidden layers of neurons, where the "hidden" term refers to layers that are not directly adjacent to the input or output. In this process, increasingly complex and abstract features can be constructed, through the addition of more layers and/or increasing the width of layers (i.e. increasing the number of neurons per layer). Correspondingly, the model can learn increasingly complex and abstract representations (i.e. "concepts" if the term is used loosely). However, for one to use more than a single hidden layer, it is necessary to determine how to assign error attribution and make corrections to its weights by working backwards originating from the predicted output, and back through the neural network. This backwards propagation of errors is known formally as "backpropagation". Although the conceptual foundation of backpropagation was discovered in 1963,[11] it was not until 1986 that Hinton and co-workers discovered a way for this algorithm to be applied to neural networks,[12] which was a key historical development that enabled practically usable ANNs.

During the process of backpropagation, an algorithm known as gradient descent is used to find the minimum in the error surface caused by each respective neuron when generating a corresponding output. Conceptually, gradient descent is no different from the steepest descent algorithm used in classical molecular dynamics simulation. The major difference is instead of iteratively minimizing an energy function and updating atomic coordinates for each step, an error function of the target output of the ANN is iteratively minimized and the weights of the neurons are updated each step, which are also known as "iteration" in the ANN literature. The data in the training set may be iterated over multiple times, with a complete pass over the data being called an "epoch."

A key issue with backpropagation is that the error signals become progressively more diffused as the signal goes back through each hidden layer. This is because, as the signal goes deeper into the model, an increasing number of neurons and weights are associated with a given error. Until recently, this made it difficult to train many layers efficiently; anything more than a few layers that required a long time to converge with a high probability of overfitting, especially for the layers closest to the output. In addition, the nonlinear transformation functions, such as sigmoids, had finite dynamic range, so error signals tends to decay as they passed through multiple layers, which is more commonly known as the "vanishing gradient problem".[13]

Since 1986, several key algorithms, including unsupervised pre-training,[14] rectified linear functions[15] and dropout,[16] have been developed to improve the training process for ANN, to address the vanishing gradient problem, and to reduce overfitting which ANN are particularly susceptible to. Perhaps the largest impediment to training deep neural networks (DNN), was the vanishing gradient problem as it practically capped the depth of the neural network. Pre-training, discovered by Hinton *et al.* in 2006 is a fast, greedy, algorithm that uses an unsupervised layer-wise approach to train a DNN one layer at a time.[14] After the pre-training phase is complete, a more subtle fine-tuning process, such as stochastic gradient descent, is used to train the model. Using the pre-training approach, the model would have already learnt the features before backpropagation begins, mitigating the vanishing gradient problem. An alternative solution emerged in 2011, where Bengio and co-workers demonstrated the rectified linear activation (ReLU) function that sidesteps the vanishing gradient problem entirely. The ReLU's first derivative is precisely unity or 0, generally ensuring that error signals can back-propagate without vanishing or exploding. (Figure 2).[15]

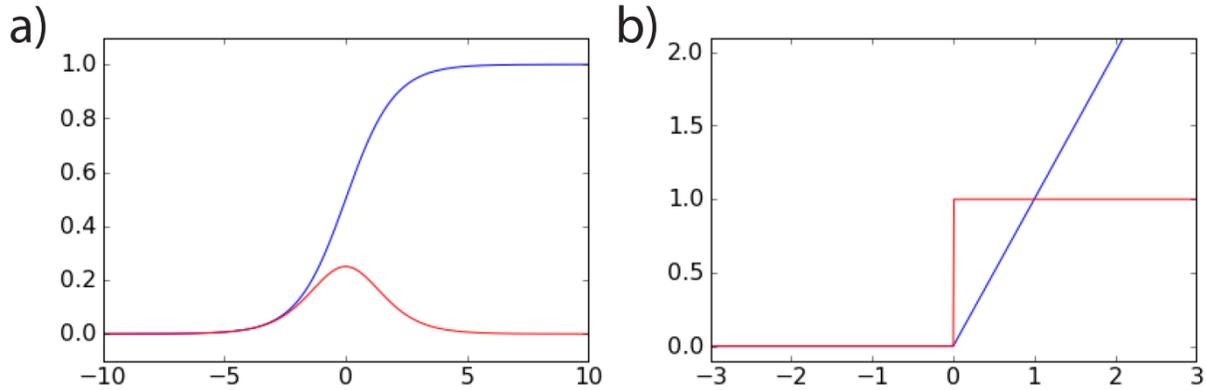

Figure 2: Plot of (a) sigmoidal and (b) rectified-linear (ReLU) function (in blue) and their corresponding first derivative (in red). Unlike the sigmoidal function, where its derivative varies with respect to the value of x, in the ReLU function, the first derivative is either 0 or 1.

As these methods enabled the training of deeper and more complex neural network architecture, overfitting also became more of an issue, which led to the development of the dropout algorithm. In dropout, for each epoch of the training process, a fixed proportion of neurons are randomly selected to be temporarily excluded from the model. The net effect of dropout is that it simulates many different architectures during training, which prevents co-dependency among neurons and reduces overfitting.[16] Whilst the architecture of modern DNNs vary widely, a popular configuration is ReLU-based neural networks. When coupled with dropout and early stopping, such ReLU networks have often been enough to regularize the model (i.e. prevent overfitting).[17]

Having provided a summary of the key developments in ANNs and its associated algorithms, we note that it is by no means comprehensive. In addition to the traditional feedforward DNN (Figure 1b) that has been discussed thus far, more recent developments include alternative architectures, notably convolutional neural networks (Figure 3a),[18,19] recurrent neural networks (Figure 3b),[19,20] and autoencoders (Figure 3c) that have been highly successful in computer vision and natural language processing applications. A technical discussion of various DNN architectures while informative to understanding the deep learning literature, is beyond the scope of this review, therefore, we refer our readers to the following prior publications summarizing this research topic.[21-24] By now, it should be evident that ANNs itself are not a new invention. Indeed, the mathematical algorithm for ANNs was developed by McCulloch and Pitts in 1943,[25] and practically trainable ANNs dates as far back to 1986, coinciding with invention of backpropagation for neural networks by Rumelhart, Hinton, and Williams.[12] Deeper neural networks beyond a few hidden layers (Figure 1b) was only achievable with more recent algorithmic developments in the last few years.[14-16] Therefore, how is DNNs not just the mere rebranding of ANNs of the last century, and how is it better than the traditional machine learning algorithms that are already successfully used in various cheminformatics applications?

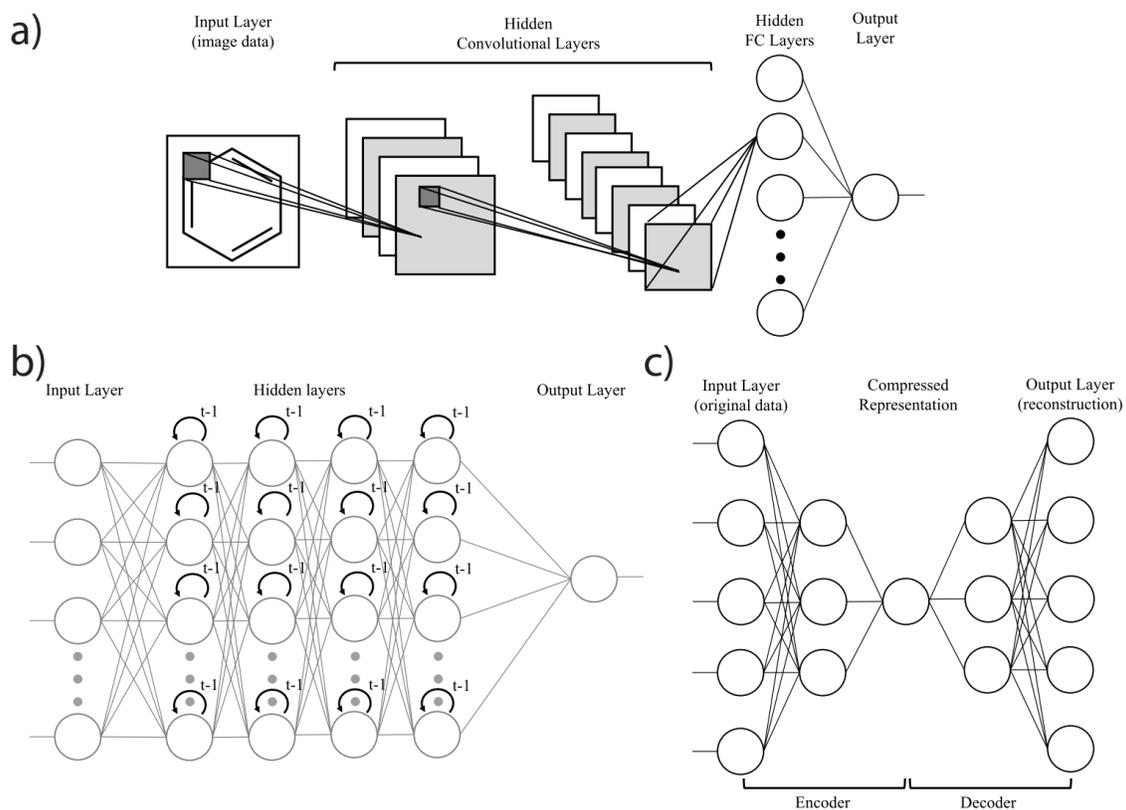

Figure 3: (a) Schematic diagram of a convolutional neural network (CNN). CNNs are designed with the explicit assumption that the input is in the form of image data. Each convolutional layer extracts and preserves the spatial information and learns a representation which is then typically passed onto a traditional fully-connected feedforward neural network before the output layer. (b) Schematic diagram of a recurrent neural network (RNN). RNNs at its simplest implementation are a modification of the standard feedforward neural network where each neuron in the hidden layer receives an additional input from the output from the previous iteration of the model, denoted as "t-1" circular arrows. (c) Schematic diagram of an autoencoder, which are neural networks used in unsupervised learning. In autoencoders, the objective is to learn the identity function of the input layer, and in the process, a compressed representation of the original data in the hidden layers is learned.

Many decades of chemistry research has led to the development of several thousand molecular descriptors that describe a range of properties of conceivably any compound. Molecular descriptors thus serve as features constructed using chemical knowledge and intuition (i.e. domain expertise) that can be used in traditional machine learning models, which have achieved reasonable success in computational chemistry applications.[26-31] Traditional machine learning algorithms such as linear regression and decision trees are intuitive and create simple models that humans can understand. Nevertheless, as we progress to the prediction of more complex properties with non-linear relationship, typically those associated with biological processes and materials engineering, it is often necessary to rely on more sophisticated and less transparent algorithms such as support vector machines (SVM) and random forests (RF) in order to achieve an acceptable level of predictive accuracy. At a first glance, deep learning algorithms falls under the latter category, but it has one major difference. Unlike SVMs and RFs, DNNs transform inputs and reconstruct them into a distributed representation across the neurons of the hidden layers. With appropriate training methods, different features will be learned by the neurons in the hidden layers of the system; this is referred to as automatic feature extraction. As each hidden layer becomes the input for the next layer of the system and non-linear transformations can be applied along the way, it creates a model that progressively "learns" increasingly abstract, hierarchical and deep features.

Automatic feature extraction, a process that requires no domain knowledge, is therefore one of the most significant benefits of a deep learning algorithm. This is unlike traditional machine learning algorithms, where a model must be carefully constructed with the "correct" features based off chemical knowledge and intuition for it to perform and generalize well. It is for this reason, that deep learning has become the dominant algorithm used in speech recognition[32] and computer vision[18,33-35] today. ImageNet is an annual assessment and competition of various algorithms for image classification. Prior to deep learning, the state-of-the-art models employed hovered in the 25-30% error rate, which falls short from the ideal goal of matching a trained human error rate of 5.1%.[36] In 2012, deep learning algorithms were first introduced to this community by Hinton and co-workers,[18] and their DNN-based model achieved a 16.4% error rate. That was a significant improvement from established models in computer vision at that time, and the second-best performing model based off traditional machine learning algorithms only achieved a 26.2% error rate. Subsequent improvements in DNN-based models eventually achieved an error rate of under 5.0%, exceeding human performance in 2015 (Figure 4), which was only 3 years after deep learning made its introduction to the computer vision field.[33,34] For practitioners in these field, the impact of deep learning and its automatic feature extraction ability has been transformative, not only in its ability to exceed "glass ceiling" expectations in the field, but the remarkably short time it has taken to achieve it. In recent years, deep learning has also demonstrated promise in other disciplines outside the computer science domain, including high-energy particle physics[37] and bioinformatics.[38]

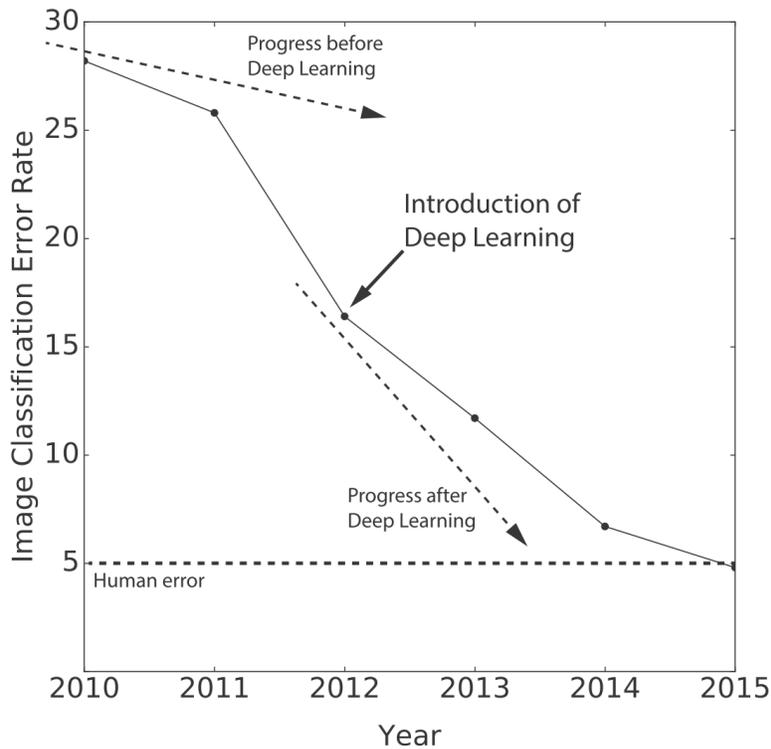

Figure 4: Historical error rate of the best performing image classification algorithms in the annual ImageNet competition.[39] Established models of computer vision stagnated at 25-30%. The introduction of deep learning in 2012 led to a significant improvement to ~15%, and human-level accuracy (~5%) for image classification was achieved by 2015.

An equally important aspect of deep learning that has not been discussed is the role of non-algorithmic developments over the years. Specifically, the availability of "big data" and the GPU hardware technological advances that were both absent in the last century have created a confluence of events that makes the advent of DNNs different from the ANNs of the last century. The seminal work in 2012 that is most widely regarded as the paper that propelled deep learning in the limelight was Hinton's AlexNet paper.[18] While algorithmic developments, notably dropout contributed to its success, the availability of a much larger dataset comprising of 1.2 million images, compared to datasets of 10,000 images used in the past, also played a critical role in its success. With the development of deeper and larger neural networks, training time can often extend to days or weeks. However, much like how the field of computational chemistry has benefited from the rise of GPU-accelerated computing,[40,41] this technology has also mitigated the training speed issues of DNNs.

Of the more practical considerations, the availability of open-source code and documentation for training neural networks on GPUs is also arguably another reason for the rapid proliferation of deep learning in recent years, including its impact on academic research as evidenced by the exponential growth of deep learning related publications since 2010 (Figure 5a). Much like how the majority of the computational chemists in modern times no longer write their own code to perform molecular dynamics simulation or run quantum chemical calculations, but instead rely on established software packages,[42-48] the deep learning research community has too reach a similar level of maturity, with the current major software packages for training neural networks including Torch, Theano, Caffe, and Tensorflow. Perhaps the oldest of the four, Torch was first released in 2002 as a machine learning scientific computing framework developed at NYU, but since then deep learning libraries has been added.[49] Theano was the first purposed-developed deep learning framework, released by Benjio and co-workers at Université de Montréal in 2008,[50] and it has since developed into a community effort with over 250 contributors. This was closely followed with the release of Caffe, developed by the Berkeley Vision and Learning Center in 2014.[51] Most recently, Tensorflow,[52] which is developed by Google was released in late 2015 has arguably gained a surge of uptake in the deep learning community, as evidenced from its spike in google search rankings (Figure 5b), and the fact that its Github has been starred and forked over 33,000 and 14,000 times respectively, despite it being only released for a little over a year. In addition, APIs, such as Keras released in 2015, has greatly simplified the construction and training of neural networks, which has significantly reduced the barrier of entry for new deep learning practitioners.

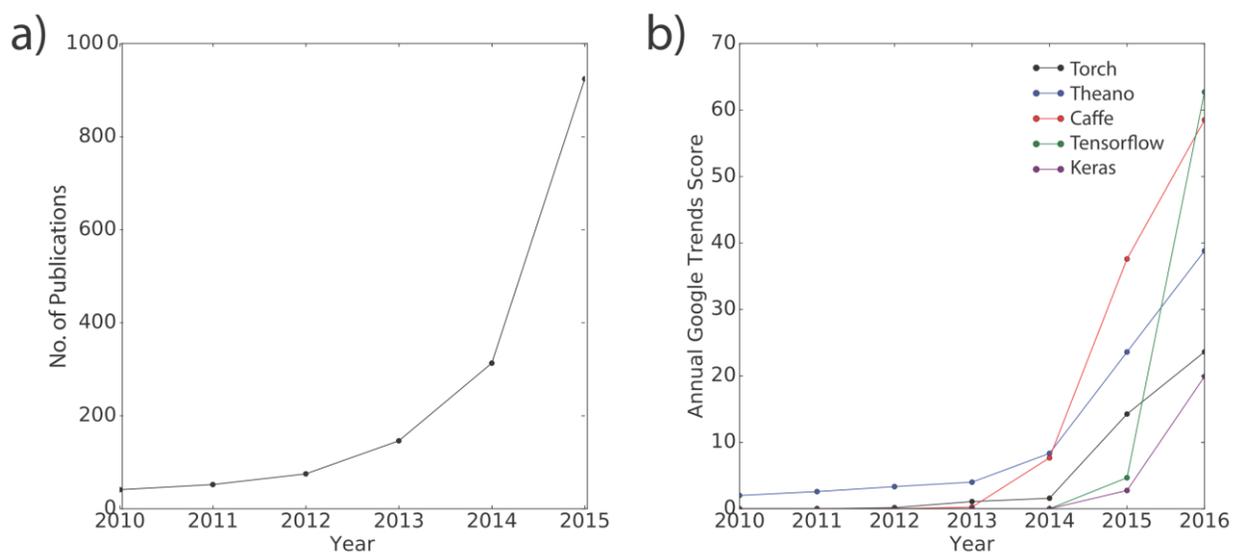

Figure 5: Growth of (a) deep learning publications as indexed by ISI, and (b) annual google trends score of major deep learning software packages since 2010.

Unquestionably, the computer science domain has been the main benefactor of the surge of mineable data obtained from the internet (Figure 6a), and not surprisingly has also been the field where deep learning had the largest impact. In chemistry, we have also seen a corresponding growth of data in publically accessible databases, such as the Protein Data Bank (Figure 6b) and PubChem (Figure 6c), with more data being generated from recent developments in high-throughput omics technologies.[53] It is for these reasons that we are optimistic that the field of computational chemistry is starting to experience the same confluence of events, and this will greatly facilitate deep learning applications in our field. We can leverage on the algorithmic breakthroughs in the computer science domain, the increasing availability of chemical data, and the now matured GPU-accelerated computing technologies (Figure 6d).

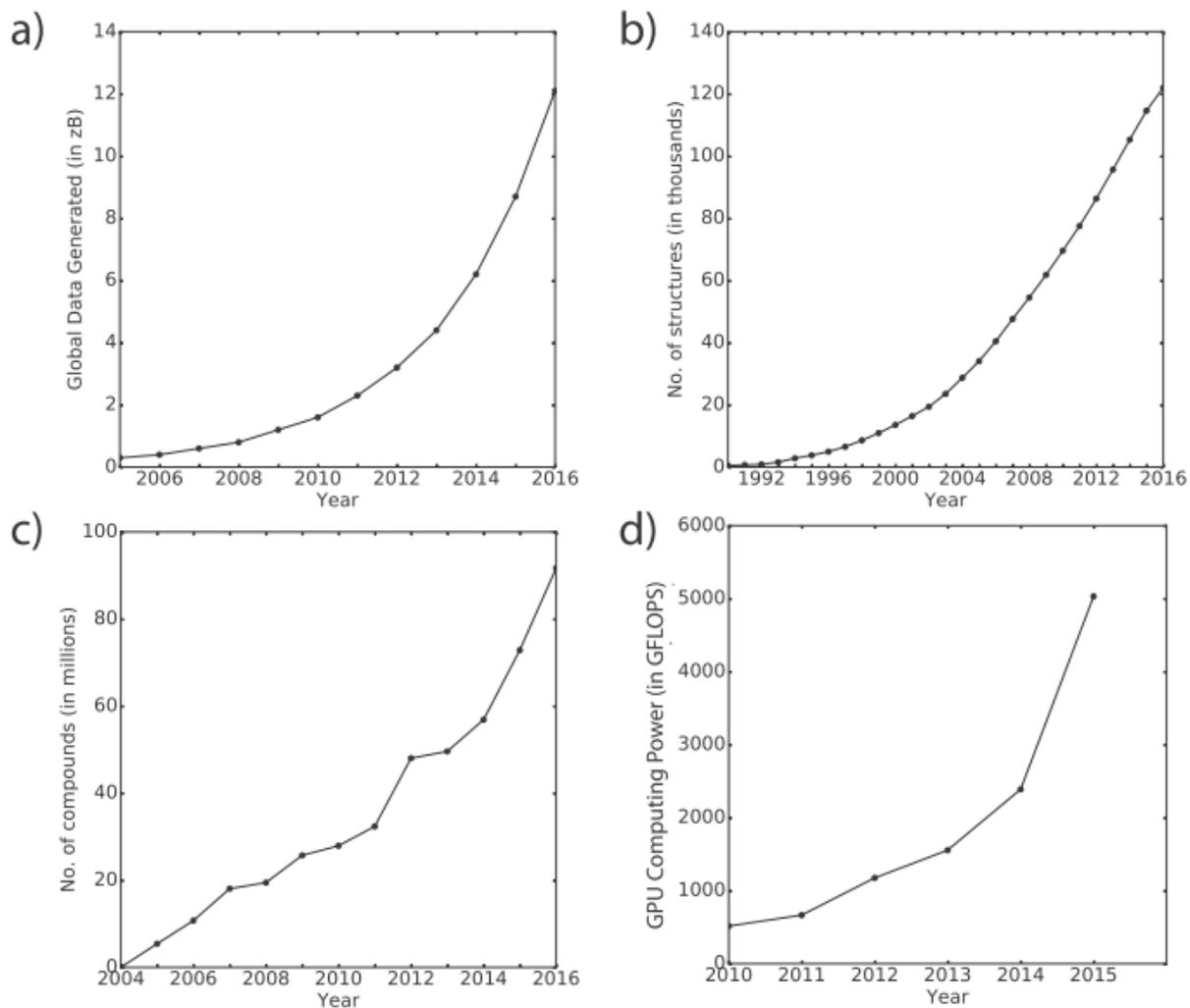

Figure 6: The growth of (a) global data generated, (b) number of structures deposited in the Protein Data Bank, (c) number of compounds deposited in PubChem and (d) GPU computing power for scientific computing,[54] all share similar parallels in their upwards trajectory.

### 3. Computer-Aided Drug Design

In computer-aided drug design, traditional machine learning algorithms have a long history in the field of cheminformatics, notably in their contribution to quantitative structure activity relationship (QSAR) applications. In QSAR, the output to be predicted is usually the biological activity of a compound. Usually regression models are used, and the input data are molecular descriptors, which are precomputed physicochemical properties of the molecule, designed from chemistry domain knowledge. Early work in QSAR applications used linear regression models, but these were quickly supplanted by Bayesian neural networks,[55-57] followed by RFs[26] and SVMs.[31] Practitioners in the field have historically favored models that allow for variable selection so that an informed chemist can determine if selected features made sense. In addition, models that allowed assessment of uncertainty of output predictions were also preferred. The field of QSAR is vast, and we refer readers to the following list of reviews for key historical technical developments.[58-61] For the purpose of this review, we will limit the scope of discussion to the performance of DNN-based QSAR models and appropriate comparisons to traditional machine learning models.

The first foray of deep learning into QSAR was the Merck challenge in 2012.[62] In this publically available challenge, teams were provided precomputed molecular descriptors for compounds and their corresponding experimentally measured activity for a total of 15 drug targets. Submitted models were evaluated on their ability to predict activity against a test set not released to participants. The winning group used DNN models, led by Dahl who was part of Hinton's research team.[62] Notably, it should be emphasized that the team had no formally trained computational chemist in the group; they were from the computer science department.

In 2014, Dahl *et, al.,* submitted an arxiv paper exploring the effectiveness of multi-task neural networks for QSAR applications, based on the algorithms used in the Merck challenge.[63] In this work, the authors used a multi-task DNN model. Here "multi-task" refers to a model that predicts not just a single output of interest, but multiple outputs simultaneously, which in their case was the results from 19 assays. The dataset used was curated from PubChem and included over 100,000 data points. Molecular descriptors totaling 3764 descriptors per molecule were generated using Dragon,[64] and they were used as the input features for the DNN. In an accuracy performance benchmark against other traditional machine learning algorithms, such as gradient-boosted decision trees and logistic regression, the DNN-based model outperformed all others in 14 of 19 assay predictions by a statistically significant margin and was comparable in terms of performance in the remaining 5 assay prediction.[63] In addition, the advantages of a multi-task neural network was noted by the authors, particularly in the fact that it develops a shared, learned feature extraction pipeline for multiple tasks. This means that not only can learning more general features produce better models, but weights in multi-task DNNs are also constrained by more data cases, sharing statistical strength.[63] Lastly, an interesting observation from that study was how DNNs were able to handle thousands of correlated input features, which goes against traditional QSAR wisdom as highlighted by Winkler in 2002,[65] although we note that the observations published by Winkler at that time was prior to the development of DNNs. In Dahl's work, the authors observed that halving the input features did not led to any performance degradation.

A subsequent study in 2015 published by Merck, comprehensively analyzed the training of DNNs and compared their performance to the current state of the art used in the field, RF-based models, on an expanded Merck challenge dataset.[66] The authors concluded that DNNs could be adopted as a practical QSAR method, and easily outperformed RF models in most cases. In terms of practical adoption, the authors emphasized the dramatic advance in GPU hardware that DNNs leverage and also the economic cost advantages of deploying GPU resources as opposed to conventional CPU clusters that are used by traditional machine learning models.[66] The key issue associated with training deep neural networks, particularly in the number of tunable parameters was also investigated. The authors discovered that most single task problems could be run on architectures with two hidden layers, using only 500-1000 neurons per layer and 75 training epochs. More complex architecture and/or longer training time yielded incremental but diminishing returns in model accuracy.[66] Despite the overall promising performance of DNNs in the Merck challenge and associated studies as summarized above, the results were received with skepticism by some practitioners in the research community.[67] Common concerns include the small sample size, and that the incremental improvements in predictive accuracy was difficult to justify in the face of increase in model complexity.

In 2014, Hochreiter and co-workers published a peer-reviewed paper at the Neural Information Processing Systems (NIPS) conference on the application of multi-task DNNs for QSAR application on a significantly larger dataset.[68] In this study, the authors curated the entire ChEMBL database, which was almost 2 orders of magnitude larger than the original Merck challenge dataset. This dataset included 743,336 compounds, approximately 13 million chemical features, and 5069 drug targets. It is also interesting that the authors did not use explicitly computed molecular descriptors as input data, but rather used ECFP4 fingerprints[69] instead. The authors benchmarked the accuracy performance of the DNN model across 1230 targets, and compared them against traditional machine learning models, including SVMs, logistic regression and others. It should be noted that gradient-boosted decision trees which performed almost as well as DNNs in Dahl's 2014 paper was not included in this study. Nevertheless, it was demonstrated that DNNs outperformed *all* models they tested on, which also included 2 commercial solutions and 3 currently implemented solutions by pharmaceutical companies (Figure 7).[68] While most traditional machine learning algorithms accuracy ranged from 0.7 to 0.8 AUC, DNNs achieved an AUC of 0.83. Of the better performing models (AUC > 0.8), DNNs also had the least severe outliers, which the authors hypothesized was due to the DNN's shared hidden representation that enabled it to predict tasks which would be difficult to solve when examined in isolation. In agreement with Dahl's 2014 study, the use of a multi-task DNN conferred two advantages: (i) allowance for multi-label information and therefore utilizing relations between tasks and (ii) allowance to share hidden unit representations among prediction tasks.[68] The authors in this study noted that the second advantage was particularly important for some drug targets where very few measurements are available, and thus suggested that a single target prediction may fail to construct an effective representation.[68] The use of multi-task DNNs partially mitigates this problem, as it can exploit representations learned across different tasks and can boost the performance on tasks with fewer training examples. Moreover, DNNs provide hierarchical representations of a compound, where higher levels represent more complex concepts that would be potentially more transferable beyond the training set data.[68]

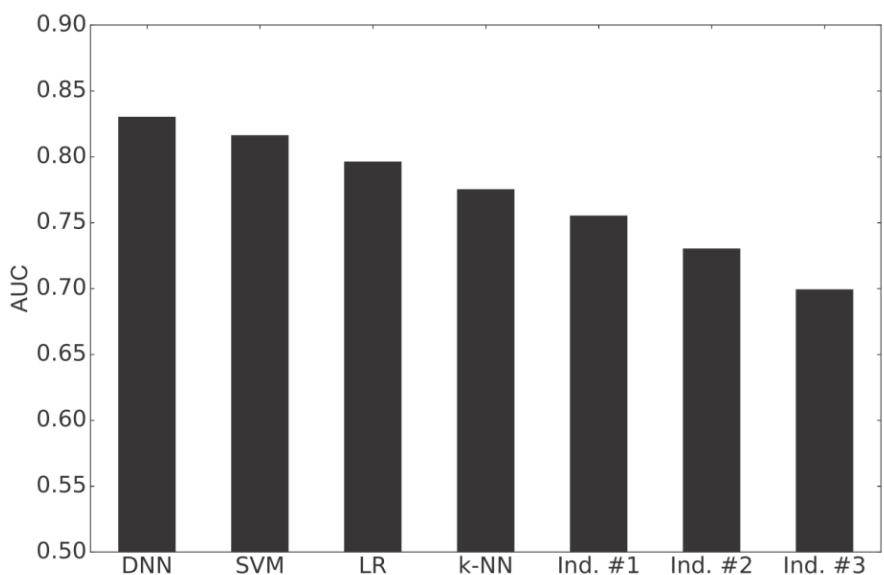

Figure 7: Performance accuracy (in terms of AUC metrics) of deep neural network against several traditional machine learning algorithms, including: support vector machines (SVM), logistic regression (LR), k-nearest neighbor (k-NN) and commercially-implemented solutions (Pipeline Pilot Bayesian Classifier, Parzen-Rosenblatt KDE-based approach and Similarity Ensemble Approach respectively) for activity prediction of a curated database obtained from ChEMBL.[68]

A similar large scale study was submitted to arxiv in 2015 by the Pande group and Google.[70] In this study, about 200 drug targets were identified, but significantly more data points (40 million) were included. Unlike the earlier NIPS paper, Pande and co-workers focused their investigation on the effectiveness of multi-task learning in DNNs rather than the performance of the DNN model itself. The authors curated a database that was combined from multiple sources of publicly available data, including PCBA from the PubChem Database,[71] MUV from 17 challenging datasets for virtual screening[72], DUD-E group[73] and the Tox21 dataset.[74] As with Hochreiter and co-workers, the molecules were featurized using ECFP fingerprints, and no explicit molecular descriptors were computed. Amongst the key findings, was that multi-task performance improvement was consistently observed, although it was not evident whether additional data or additional tasks had a larger effect in improving performance.[70] The authors also observed limited transferability to tasks not contained in the training set, but the effect was not universal and required large amounts of data when it did work successfully, which partially reinforces the claims of multi-task learning advantages proposed by Hochreiter and Dahl.[63,68] Curiously, the multi-task improvement varied in degree from one dataset to another, and no satisfactory explanation was provided. Nevertheless, the consistent outperformance of multi-task DNNs against traditional machine learning models such as logistic regression and RF was evident (Figure 8), where the performance lift in AUC ranges from 0.02 to 0.09.[70]

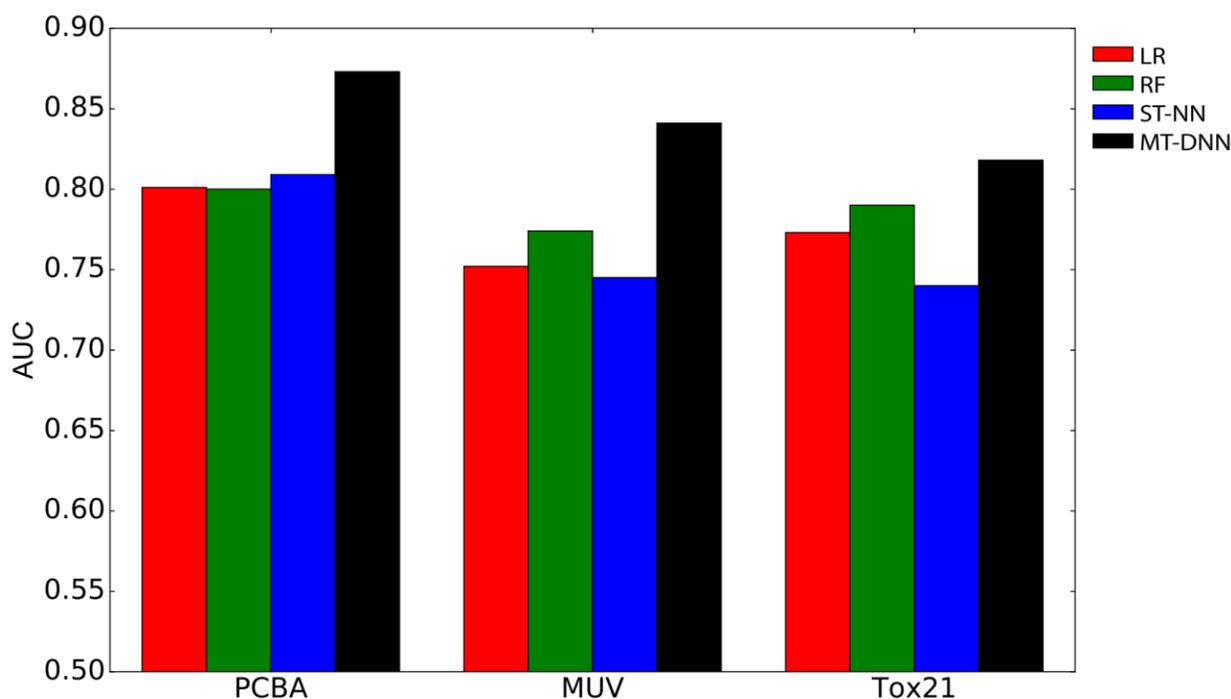

Figure 8: Consistent performance lift in accuracy (in terms of AUC metrics) was observed across 3 different databases (PCBA, MUV, Tox21) when using multi-task deep neural networks (MT-DNN) as compared to logistic regression (LR), random forest (RF) and single-task neural network (ST-NN).[70]

To date, there has been at least 4 reported applications of DNNs for QSAR, with consistent observations that deep learning outperforms traditional machine learning counterparts. However, all of the studies have thus far mostly focused on biological activity prediction. Conceptually, DNNs should have similar performance in predicting other properties of interest, which may include ADMET properties, as well as applications in other parts of computer-aided drug design, such as in virtual screening.

Drug-induced liver injury (DILI) is the most frequent cause of safety-related drug withdrawals over the last 5 decades.[75] Mechanisms underlying DILI are complicated and diverse, drugs that cause DILI in humans are not easily probed by conventional means, making toxicological studies of DILI difficult. A recent study from Xu et. al. used DNNs to predict DILI toxicity.[76] The authors used both explicit molecular descriptors, computed from Mold[77] and PaDEL,[78] as well as the URGNN method for molecular structuring encoding developed by Lusci et. al.[79] as input data for the DNNs. The model was trained on 475 drugs, with an external test set of 198 drugs, and the best model that utilized a DNN had accuracy of 86.9%, sensitivity of 82.5%, specificity of 92.9%, and AUC of 0.955.[76] In comparison, traditional models have lower absolute performance metrics by 10-20%.[76] Interestingly, using input from the URGNN molecular structural encoding method, the authors created a model with the highest performance (AUC 0.955), outperforming similarly trained DNNs that used calculated molecular descriptors from Mold (AUC 0.931) and PaDEL (AUC 0.895).[76] This suggests that a good molecular encoding method such as UGRNN may be more effective in providing the necessary features to DNNs, as deep learning algorithms have the ability to automatically extract the necessary features, and this ability may be on par or perhaps even better than domain-expert feature engineering through the development of explicit molecular descriptors.

Another application for DNN modeling toxicity was published by Swamidass and co-workers in 2015.[80] One common mechanism of drug toxicity stems from electrophilic reactive metabolites that covalently bind to proteins. Epixodes are a functional group of this nature, which are often formed by cytochrome P450 metabolism of drug molecules, which acts on aromatic or double bonds. Swamidass and co-workers results were particularly distinctive, because they developed a DNN model to predict the specific location on a molecule that undergoes epoxidation, i.e. its site of epoxidation (SOE). This work was based off an earlier model, Xenosite, an ANN-based model for P450 metabolism on small-molecules, which despite being a shallow network, was already outperforming the accuracy of SVM-based models by as much as 5%.[81] Further improvements were subsequently achieved by investigating the effect of using different types of molecular fingerprints for modeling P450 metabolism, where they discovered that further accuracy gains can be achieved by using a consensus model utilizing different fingerprint types,[82] and a related sister model that predicted the site of glucoronidation metabolism.[83] In their more recent work on predicting epoxide-based toxicity, Swamidass and co-workers designed a 4-layer DNN architecture, and trained the model on a database of 702 epoxidation reactions, and identified SOEs with 94.9% AUC performance, and separated (i.e. classified) epoxidized and non-epoxidized molecules with 79.3% AUC.[80] Moreover, within epoxidized molecules, the model was able to provide atomic-level precise information, by separating aromatic or double bond SOEs from all other aromatic or double bonds with AUCs of 92.5% and 95.1%, respectively.[80] This makes the DNN model the first mechanistic model in the literature, which not only predicts the formation of reactive epoxides of drug candidates, but also accurately identifies the specific epoxidized bonds in the molecule. Using a similar DNN model, Swamidass and co-workers modelled the site of reactivity of small-molecules towards soft nucleophiles such as gluthaione (GSH).[84] By training only on qualitative reactivity data, they were able to construct a DNN-based model that identified sites of reactivity within reactive molecules with 90.8% accuracy, and separate reactive and unreactive molecules with 80.6% accuracy.[84] In addition, the model's predictions correlated well with quantitative GSH reactivity measurements in external data sets that were more chemically diverse, indicating the model's generalizability across a larger area of chemical space.[84] A subsequent publication expanded the scope of the model to encompass reactivity towards GSH, cyanide, protein, and DNA. The resulting model yielded a cross-validated AUC performance of 89.8% for DNA and 94.4% for protein, and separated electrophilically reactive molecules with DNA and protein from nonreactive molecules with a cross-validated AUC performance of 78.7% and 79.8%, respectively.[85] Furthermore, the model's performance also significantly outperformed reactivity indices calculated from QM methods.[85] As drug toxicity is often caused by electrophilic reactive metabolites, models that assist in the study of identifying site reactivity, which has been up to now conspicuously absent in the literature, can potentially be utilized to construct a mechanism-based prediction of molecule toxicity.

A larger scale study on chemical toxicity was also recently published by the Hochreiter group in 2016.[86] In this work, the authors reported on the application of DNN models on the Tox21 Data Challenge released by NIH in 2014. The database consisted of 12,000 environmental chemicals and drugs, and their corresponding measurements on 12 different assays designed to measure a variety of toxicity effects. Not surprisingly, the DeepTox model developed by Hochreiter and co-workers had the highest performance of all methods submitted to the Tox21 challenge.[9] Further analysis of their model indicated that using a multi-task DNN model led to consistent outperformance against single-task models in 10 out of 12 assay predictions.[86] Additional benchmarks to traditional machine learning algorithms, including SVM, RF and Elastic

Net, also demonstrated that DNN outperformed in 10 out of 15 cases.[86] Lastly, while the original DeepTox model used molecular descriptors provided by NIH in the Tox21 challenge, the authors also showed that a similarly trained DNN model developed using only ECFP4 fingerprint as input data had similar performance to those trained on explicit molecular descriptors, which is similar to the observations made by Xu et. al. in their DILI toxicity model.[76] Interestingly, on visualization of the first hidden layer of these DNNs, the author observed that 99% of neurons in that layer had a significant association with at least one known toxicophore feature, suggesting that deep learning can possibly support the discovery of new chemical knowledge in its hidden layers.[86]

In line with the progress in QSAR and toxicity prediction, deep learning algorithms have also started to make an impact in other aspects of computer-aided drug design. In 2013, Baldi and co-workers reported using a DNN model to predict molecule solubility.[79] More recent research developments in this direction was also submitted to arxiv by Pande and co-workers, where they developed a multi-task DNN model for predicting not just solubility, but the entire spectrum of ADMET properties.[87] Deep learning may also have a future in virtual screening as a viable alternative or complement to existing docking methods. In 2016, an arxiv paper was submitted by AtomNet, a startup that developed a DNN model to classify the activity of small molecules docked in protein binding pockets.[88] Remarkably, the AtomNet DNN model was able to achieve AUC metrics ranging between 0.7 to 0.9 depending on the test set used, which significantly outperforms conventional docking methods, specifically Smina,[89] a fork of AutoDock Vina[90] by 0.1 to 0.2.[88] For additional recent developments of deep learning in applications that are more closely aligned to computational biology, we refer our readers to the following reviews that focuses on that research topic.[91]

## 4. Computational Structural Biology

Predicting the spatial proximity of any two residues of a protein sequence when it is folded in its 3D structure is known as protein contact prediction. The prediction of contacts between sequentially distinct residues thus imposes strong constraints on its 3D structure, making it particularly useful for *ab initio* protein structure prediction or engineering. While the use of physics-based simulation methods, such as long-timescale molecular dynamics[92,93] can be used for *ab initio* protein structure prediction, the computational demands are formidable. Complementary methods, such as knowledge based physical approaches developed by the groups of Wolynes, Onuchic and others are also an option,[94,95] although their computational expense while lower is still sufficiently demanding that it cannot be used for large-scale studies. Therefore, machine learning approaches are viable alternatives, including those based off ANNs,[96-98] SVMs,[27] and hidden Markov model.[99] Other approaches include template-based approaches that use homology or threading methods to identify structurally similar templates to base an inference of protein contact prediction.[100,101] The assessment of these various models for contact predictors is one of the highlights of the Critical Assessment of protein Structure Prediction (CASP) challenge which started in 1996. Despite improvements over the years, the long-range contact prediction has historically hit a glass ceiling of just below 30% accuracy. The key historical developments of computational protein structure prediction is voluminous, and we refer interested readers to existing reviews on this topic.[102-105] For the purpose of this review, we will limit the scope of discussion to the performance of recent DNN-based models, and how they have been critical to breaching the historical glass ceiling expectations in the field.

In 2012, Baldi and co-workers developed CMAPpro, a multi-stage machine learning approach, which improved contact prediction accuracy to 36%.[106] Three specific improvements were implemented in CMAPpro over earlier models. The first is the use of a 2D recursive neural network to predict coarse contacts and orientations between secondary structure elements. In addition, a novel energy-based neural network approach was used to refine the prediction from the first network and used to predict residue-residue contact probabilities. Lastly, a DNN architecture was used to tune the prediction of all the residue–residue contact probabilities by integrating spatial and temporal information.[106] CMAPpro was trained on a 2356-member training set derived from the ASTRAL database.[107] For cross-validation purposes, the set was segmented into 10 disjoint groups belonging to different SCOP fold, which meant that neither training nor validation set shared sequence or structural similarity. The resulting model performance was then tested against a 364-member test set of new protein folds reported between version 1.73 and 1.75 release of the ASTRAL database. CMAPpro performance was compared against several permutations of the multi-stage machine learning model, including a single hidden layer neural network (NN), a single hidden layer neural network that utilized the coarse contact/orientation and alignment predictors, which is generated by the 2D recursive neural network and the energy-based neural network (NN+CA), and a deep neural network but without CA features (DNN). Based on the relative performance, both the deep network architecture and CA features were required to achieve an accuracy of 36%; DNN and NN+CA each achieved 32%, while NN which represents the previous state-of-the-art only achieved 26% accuracy.[107]

A different implementation of DNN for protein contact prediction was also reported by Eickholt and Cheng in 2012.[108] In their algorithm, DNCON, it combined deep learning with boosting techniques that was used to develop an ensemble predictor. A 1426-member dataset derived from the Protein Data Bank was used to train DNCON, with a random split between the training (1230-member) and validation (196-member) set. Explicitly engineered features were used as input for the DNN. Specifically, three classes of features were used: (i) those from two windows centered on the residue pair in question (e.g. predicted secondary structure and solvent accessibility, information and likelihoods from the PSSM and Acthley factors, etc.), (ii) pairwise features, (e.g. Levitt's contact potential, Jernigan's pairwise potential, etc.) and (iii) global features (e.g. protein length, percentage of predicted exposed alpha helix and beta sheet residues, etc.).[108] Using these engineered features, the DNN model was tasked to predict whether or not a particular residue pair was in contact. In addition, boosted ensembles of classifiers was created by training several different DNNs using a sample of 90,000 long-range residue-residue pairs from a larger pool obtained from the training set. In evaluating its performance, cross-validated accuracy of DNCON was 34.1%. The model's performance transferability was demonstrated in its performance benchmarks against the two best predictors of CASP9,[109] ProC_S3[28] and SVMcon,[27] which are based off RF and SVM algorithms respectively. In that assessment, the respective test set was used for each software. While the improvement was not as dramatic as that reported by Baldi and co-workers, DNCON performance was ~3% better than the state-of-the-art algorithms for its time; ProC_S3 (32.6% vs 29.7%) and SVMcon (32.9% vs 28.5%).[108]

Both DNN-based protein contact prediction models were noteworthy, as it enabled the community to breakthrough the 30% accuracy barrier that was not possible in prior years. Apart from protein contact prediction, DNNs have also been successfully applied to the prediction of various protein angles, dihedrals, and secondary structure from only sequence data. Using DNNs, Zhou, Yang and co-workers published a series of sequence-based predictions for Cα-based angles and torsions.[110-112] Unlike protein contact prediction, backbone torsions are arguably better restraints for use in *ab initio* protein structure prediction and other modeling purposes.[113] In the development of these DNN-based models, Zhou, Yang and co-workers used a 4590-member training set and a 1199 independent test set obtained from the protein sequence culling server PISCES.[114] Input data included specifically engineered features obtained from the Position Specific Scoring Matrix generated by PSI-BLAST,[115,116] as well as several other physicochemical properties related to residue identity, including steric, hydrophobicity, volume, polarizability, isoelectric point, helix probability, amongst others.[117]

In the development of the SPINE-X algorithm, a DNN was used to predict secondary structure, residual solvent-accessible surface area (ASA), φ and ψ torsions directly.[111] A six-step machine learning architecture was developed where outputs such as ASA were used as subsequent inputs for other properties to be predicted, such as the torsions. Based on the evaluation of the model's performance on the independent test set, it achieved a mean absolute error of $22^0$ and $33^0$ respectively for the φ and ψ dihedrals. Secondary structure prediction accuracy on independent datasets were ranging from 81.3% to 82.3%, and this achievement is noteworthy, considering that the field of secondary structure prediction from sequence data has stagnated just under 80% accuracy in the recent decade, some of which utilize traditional machine learning algorithms.[118] In a similar fashion, for the SPIDER algorithm that was developed later, a DNN was used to predict Cα angles (θ) and torsions (τ) directly.[110] Based on the evaluation of the model's performance, it achieved a mean absolute error of $9^0$ and $34^0$ for θ and τ respectively, and the authors observed that the model's error increased from helical residues to sheet residues to coil residues, following the trend in unstructuredness. Using these predicted angles and torsions as restraints, the authors were able to model the 3D structure of the proteins with an average RMSD of 1.9A between the predicted and native structure.[110] The SPINE-X and SPIDER algorithm was subsequently re-trained as a parallel multi-step algorithm that predicted simultaneously the following properties: secondary structure, ASA, φ, ψ, θ and τ.[112] This resulted in a modest improvement in overall accuracy of secondary structure by 2%, and reduction of MAE by 1-3$^0$ for the angles/torsions, while maintaining the same level of ASA performance.

Apart from protein structure modeling, deep learning has also been utilized to predict other properties of interest based on sequence data. For example, predicting sequence specificities for DNA and RNA-binding proteins was recently reported.[119,120] In the seminal work by Frey and co-workers,[119] the DeepBind algorithm was developed to predict the sequence specificities of DNA and RNA-binding proteins. Using 12 terabases of sequence data, spanning thousands of public PBM, RNAcompete, ChIP-seq and HT-SELEX experiments, the raw data was used as an input into a DNN algorithm to compute a predictive binding score. DeepBind's ability to characterize DNA-binding protein specificity was demonstrated on the PBM data from the revised DREAM5 TF-DNA Motif Recognition Challenge by Weirauch et. al.[121] Notably, DeepBind outperformed *all* existing 26 algorithms based on Pearson correlations and AUC metrics, and was ranked first amongst 15 teams in the DREAM5 submission.[119] Interestingly, their results also indicated that models trained on *in vitro* data worked well at scoring *in vivo* data, suggesting that the DNNs has captured a subset of the properties of nucleic binding itself.

As with the repeated occurrence of deep learning outperforming traditional machine learning algorithms in other fields,[18,32-35] as well as in computer aided drug design itself,[63,68,70] the utilization of DNNs in pushing the "glass ceiling" boundaries of protein contact prediction and secondary structure prediction should come as no surprise. Conspicuously absent from this review is the application of deep learning for RNA structure prediction and modeling, which to the best of our knowledge has yet to be reported. Compared to the protein database, available structural data on RNA is smaller. Furthermore, most RNA structural data are not crystallographic but are instead NMR-based, which itself is subjected to a higher uncertainty by virtue of the fact that NMR-structures themselves are approximations resolved using physics-based force field against experimentally bounded restraints.[122] Nevertheless, it will be interesting to see how deep learning can benefit the RNA modeling community.

Lastly, an interesting contrast in the use of deep learning in computational structural biology applications compared to computer-aided drug design, is the exclusive use of engineered features, and for some cases, the engineering of the architecture of the multi-stage machine learning algorithm itself. While the findings from the computer-aided drug design field is preliminary, there are some indications that explicitly engineered features do not necessarily perform better against chemical fingerprints, which arguably require less chemical domain knowledge to construct. While we concede that proteins are considerably more complex than small molecules, it would be interesting to determine if the performance of DNN models that uses input data that includes only basic structural and connectivity information, without any specifically engineered features, can accurately predict properties such as protein secondary structure, and long-range contacts.

5.  **Quantum Chemistry**

Using machine learning to supplement or replace traditional quantum mechanical (QM) calculations has been emerging in the last few years. In this section, we will examine some machine learning applications to quantum chemistry, and examine the relative performance of similar DNN-based models. In 2012, von Lilienfeld and co-workers developed a machine learning algorithm based on non-linear statistical regression to predict the atomization energies of organic molecules.[29] This model used a 7000-member subset of the molecular generated database (GDB), a library of $10^9$ stable and synthetically-tractable organic compounds. The target data used for training was atomization energies of the 7000 compounds calculated using the PBE0 hybrid functional. No explicit molecular descriptors were used as input data, instead only the Cartesian coordinates and nuclear charge were used in a "Coulomb" matrix representation. Arguably, without explicitly engineered features, this type of representation in the input data would be of the same level as that provided by molecular fingerprints used in classical molecular modeling approaches. Using only 1000 compounds for the training set, von Lilienfeld and co-workers achieved a mean absolute error (MAE) accuracy of 14.9 kcal/mol. Further tests on an external 6000 compound validation set yield similar accuracy of 15.3 kcal/mol, demonstrating the transferability of the model within "in class" compounds. What was particularly groundbreaking about this work was the ability to reasonably recapitulate QM-calculated energies, with a mean-absolute error of ~15 kcal/mol, without having any implementation of the Schrodinger Equation in the machine learning algorithm at all. More importantly, considering that this work used a traditional machine learning algorithm that lacks the advantages of DNN, and based on DNN's historical performance, it suggests that a DNN-based model should perform even better.

A subsequent publication by Hansen *et. al.* investigated a number of established machine learning algorithms, and the influence of molecular representation on the performance of atomization energy predictions on the same dataset as used in von Lilienfeld work.[123] Amongst the key findings was that using a randomized variant of the 'Coulomb matrix' greatly improved the accuracy of atomization energies to achieve as low a MAE as 3.0 kcal/mol.[123] Apart from being an inverse atom-distance matrix representation of the molecule, the randomized variant is unique and retains invariance with respect to molecular translation and rotation. An added "side effect" of this improved representation was that it was the richest one developed, as it is both high-dimensional and accounting for multiple indexing of atoms.[123] The authors discovered that sorting various representation by information did yield a correspondingly lower accuracy across all machine learning algorithms tested,[123] which highlighted the importance of good data representation in QM applications. In fairness, it should also be noted that the authors did benchmark ANNs, and while they performed satisfactorily with a MAE of 3.5 kcal/mol, it was not considerably better than non-linear regression methods of MAE of 3.0 kcal/mol. Nevertheless, we highlight the neural network used was "shallow" with a few layers, and together with the lack of a larger dataset, does not represent a true DNN implementation. One particularly illuminating conjecture from this paper is by extrapolating the performance (MAE error) with respect to the size of the dataset used, the authors concluded that 3 kcal/mol was probably the "baseline" error that one could achieve regardless of the machine learning algorithm used.[123]

In 2013, von Lilienfeld reported the application of the first multi-task DNN model that not only predicted atomization energies, but several other electronic ground and excited state properties.[124] In this work, they attempted to capitalize on the advantages of multi-task learning, by predicting several electronic properties and potentially capturing correlations between seemingly unrelated properties and levels of theory. The data was represented using the randomized variant of the 'Coulomb matrix'.[123] The target data was atomization energies, static polarizabilities, frontier orbital eigenvalues HOMO and LUMO, ionization potential and electron affinity calculated using several different level of theory such as PBE0, ZINDO, GW and SCS. The atomization energy maintained a similar accuracy of MAE of 0.16 eV (~3.6 kcal/mol) and achieved comparable accuracy of MAE of 0.11 to 0.17eV (~2.5 to 3.9 kcal/mol) for the other energy predictions, including HOMO, LUMO, ionization potential and electron affinity.[124] Furthermore, this level of accuracy was similar to the error of the corresponding level of theory used in QM calculations for constructing the training set.

While using machine learning algorithms to replace QM calculations is enticing, an alternative more "first principles grounded" approach is to use machine learning algorithms to supplement existing QM algorithms. As first reported by von Lilienfeld and co-workers in 2015, they demonstrated the Δ-learning approach, whereby a machine learning "correction term" was developed.[125] In that study, the authors used DFT calculated properties and were able to predict the corresponding quantity at the G4MP2 level of theory using the Δ-learning correction term. This composite QM/ML approach combines approximate but fast legacy QM approximations with modern big-data based QM estimates trained on expensive and accurate results across chemical space.[125] However, we noted that this approach has thus far been only demonstrated using traditional machine learning algorithms. If the performance boost using multi-task DNNs that we have observed on numerous instances applies to this example, a DNN-based approach would potentially yield superior results, but that has yet to be reported in the literature.

To the best of our knowledge, the fewer examples of DNN in quantum chemistry applications seem to indicate that it is in an earlier stage of development compared to computer-aided drug design and computational structural biology. From the literature, we know that traditional machine learning models have been used in other QM applications, such as modeling electronic quantum transport,[126] learning parameters for accurate semi-empirical quantum chemical calculations,[127] etc. In addition, new representation and fingerprints for QM applications are also being developed.[128,129] Given the observed superior accuracy of DNN-based models against traditional machine learning models in other fields of computational chemistry, we suggest that the development of DNN-based model for these classical examples of machine learning QM applications would be beneficial for the field.

**6.     Computational Material Design**

The logical extension of DNN applications in the field of quantum chemistry is to predict and design material properties that are correlated to or based on QM-calculated properties. Quantitative structure property relationship (QSPR), which is the analogous version of QSAR in the non-biological domain, is the science of predicting physical properties from more basic physiochemical characteristics of the compound, and it has been extensively reviewed in prior publications.[130,131] Similar to the early years of modern drug development, material discovery is primarily driven by serendipity and institutional memory.[132] This has relegated the field to exploratory trial-and-error experimental approaches, and the key bottleneck in molecular materials design is the experimental synthesis and characterization. In recent years, the paradigm of computational and rational materials design has been encapsulated under the materials genome initiative.[133,134] Due to the newness of this field, in this section, we will examine a few key accomplishments of using machine learning for computational material design, and highlighting deep learning applications where available.

A recent high profile example of using machine learning models to accelerate materials property research was published by Raccuglia *et. al.* in 2016.[30] The synthesis of inorganic-organic hybrid materials, such as metal organic frameworks (MOFs), have been extensively studied for decades, but the theoretical understanding of the formation of these compounds are only partially understood. In the work by Raccuglia *et. al.*, the authors used a SVM-based model to predict the reaction outcomes for the crystallization of templated vanadium selenites. What was interesting about their work, was the inclusion of "dark" reactions in training the model, which are failed or unsuccessful reactions collected from archived laboratory notebooks. The resulting model had an 89% success rate, as defined by the synthesis of the target compound type. Notably, this exceeded the human intuition success rate of 78%.[132] While a DNN-based model was not used in the study *per se*, there is no technical reason why it could not be used in place of SVM as a tool used for computational materials synthesis prediction.

One example of how DNN has been used to accelerate materials discovery was reported by Aspuru-Gizik and co-workers in 2015.[135] Here, the authors used the dataset obtained from the Harvard Clean Energy Project – a high-throughput virtual screening effort for the discovery of high-performance organic photovoltaic materials. The metric to be predicted is power conversion efficiency (PCE) which is a function of the HOMO an LUMO energies and several other empirical parameters.[135] As no high quality 3D data was available to generate Coulomb matrices, the authors decided to use fingerprints based on molecular graphs as input representation. Four different representations were tested and the results showed generally consistent accuracy (within the same order of magnitude) across HOMO, LUMO and PCE predictions. The dataset consisted of 2,000,000 compounds randomly selected from the CEPDB database and another 50,000 was extracted as the test set. Testing errors of HOMO and LUMO was 0.15 and 0.12eV respectively which was almost a 5-fold improvement relative to prior non-neural network machine learning algorithms.[135]

While DNN applications in material design is still at its infancy, it would be interesting to see how its application will fare against traditional QSPR applications and upcoming rational materials design endeavors, such as in the prediction of spectral properties of fluorophores,[136,137] properties of ionic liquids,[138] and nanostructure activity.[139]

7.  **Reservations about Deep Learning and Of Being a Black Box**

Machine learning algorithms, while they may not be the first tool of choice for many practitioners in our field, undeniably possess a rich history in the cheminformatics field and in applications like QSAR and protein structure prediction. While it may be argued that deep learning in some sense is a resurgence of the previous artificial neural network, the algorithmic and technological breakthroughs in the last decade has enabled the development of staggeringly complex deep neural networks, allowing training of networks with hundreds of millions of weights. Coupled with the growth of data and GPU-accelerated scientific computing, deep learning has overturned many applications in computer science domains, such as in speech recognition and computer vision. Given the similar parallels in the chemistry world, it suggest that deep learning may be a valued tool to be added to the computational chemistry toolbox. As summarized in Table 1, which presents key preliminary publications of DNN-based models, we have noted the broad application of deep learning in many sub-fields of computational chemistry. In addition, the performance of DNN-based model is almost always equivalent to existing state-of-the-art non neural-network models, and at times provided superior performance. Nevertheless, we have noticed that the performance lift in many cases are not as significant, if one is to make a comparison to the improvements DNN has brought to its "parent" field of speech recognition and computer vision. One mitigating factor that explains the lack of a revolutionary advance in chemistry could be the relative scarcity of data. Unlike the computer science domain where data is cheap, especially when obtained from the internet or social media, the quantity of usable data in chemistry is understandably smaller and more expensive since actual experiments or computations are needed to generate useful data. In addition, the field of chemistry has been around for centuries and given the fact that chemical principles are based on the laws of physics, it is not unconceivable that the development of features such as molecular descriptors to explain compound solubility for example, would be an easier task than developing features to explain the difference between a dog and a cat, a common task in computer vision. Therefore, with more accurate and better engineered features in chemistry, it is also plausible that we might not see such a large initial performance improvement, especially for the relatively simpler chemical principles or concepts.

Table 1: Meta-analysis of DNN-based model performance relative to state-of-the-art non-DNN models in various computational chemistry applications. Only appropriate comparisons are summarized; models trained on similar/identical datasets, using either information extracted from publications by the same group that reported multiple ML models or publically available competition.

| Prediction / Competition | DNN Models | Comments | Non-DNN Models | Comments |
|---|---|---|---|---|
| Merck Kaggle Challenge (Activity) | 0.494 $R^2$ | DNN-based model was the top performing model in the competition.[62] | 0.488 $R^2$ | Best non-DNN model in the competition.[140] |
| | 0.465 $R^2$ | Median DNN-based model recreated by Merck post-competition.[66] | 0.423 $R^2$ | Best non-DNN model (RF-based) by Merck post-competition.[66] |
| Activity | 0.830 AUC | MT-DNN based model trained on the ChEMBL database.[68] | 0.816 AUC | Best non-DNN model (SVM) trained on the ChEMBL database.[68] |
| | 0.873 AUC | MT-DNN based model trained on the PCBA database.[70] | 0.800 AUC | Best non-DNN model (RF) based model trained on the PCBA database.[70] |
| | 0.841 AUC | MT-DNN based model trained on the MUV database.[70] | 0.774 AUC | Best non-DNN model (RF) based model trained on the MUV database.[70] |
| NIH Tox21 Challenge (Toxicity) | 0.846 AUC | DeepTox (MT-DNN based model) was the top performing model.[86] | 0.824 AUC | Best non-DNN model (multi-tree ensemble model) was placed 3rd in the Tox21 challenge.[141] |
| | 0.838 AUC | Runner up in Tox21 challenge was based off associative neural networks (ASNN).[142] | | |
| | 0.818 AUC | Post-competition MT-DNN model.[70] | 0.790 AUC | Post-competition RF model.[70] |
| Atom-level Reactivity/ Toxicity | 0.949 AUC | DNN-based model that predicts site of epoxidation, a proxy for toxicity.[80] | - | No comparable model in the literature that can identify site of reactivity or toxicity. |
| | 0.898 AUC | DNN-based model that predicts site of reactivity to DNA.[84] | | |
| | 0.944 AUC | DNN-based model that predicts site of reactivity to protein.[84] | | |
| Protein Contact | 36.0% acc. | CMAPpro (DNN-based model).[106] | | |
| | 34.1% acc. | DNCON (DNN-based model).[108] | 29.7% acc. 28.5% acc. | Best non-DNN model reported in CASP9, ProC_S3 (RF-based model)[28] and SVMcon (SVM-based model)[27] are listed respectively. |

Furthermore, as computational chemists, there is a greater emphasis placed on conceptual understanding compared to engineers or technologists which is arguably the more prevalent mindset in the computer science field. In this regard, deep learning algorithms currently fall short on two accounts. First, it lacks the conceptual elegance of a first principles model that is based on the actual laws of physics, and second, DNNs are essentially a black box; it is difficult to understand what the neural network has "learned" or exactly how it is predicting the property of interest.

To address the first issue of conceptual elegance, from a certain perspective, this objection may be more of a philosophical argument of scientific preferences. In most computational chemistry applications, unless one is solving the Schrodinger Equation exactly, which we know is impossible for anything but a two body system, one must make approximations to the model. In that sense, almost all of computational chemistry is an empirically-determined, and at times even intuitively-determined, approximation of the "true" first principles Schrodinger Equation. To illustrate this point, let us examine the historical development of classical molecular modeling force fields, such as CHARMM[42] and AMBER.[43] For example, the parameterization of dihedral angle force constants have historically been targeted to QM-calculated values, the "true" values grounded in validated physical principles. However, because the dynamics of real molecules do not behave in an additive fashion (which itself is another approximation that classical molecular modeling makes), more recent re-parameterization have started modifying dihedral parameters to empirically fit experimental NMR distribution, even though that may lead to deviations from the QM-calculated values.[143,144] Similarly, the choice of columbic interactions to model electrostatic forces is only approximately correct, and recent parameter development of modeling charged ion interactions have started fitting to various experimental observables such as osmotic pressure values, and the introduction of non-physical correction terms when modeling specific pairs of electrostatic interactions.[145-147] In these examples, approximation from first principles have to be made, and this process is a human decision that is based on empirical data or at times "chemical intuition" – which as Raccuglia *et. al.* have shown, is not infallible and not always more accurate.[132] At the risk of oversimplification of the work that computational chemist do, the development of existing computational chemistry models may be viewed as an elaborate curve fitting exercise. Instead of using human expert knowledge, a conceivable alternative may be to use a deep learning algorithm to "suggest", or perhaps even help us "decide" what approximations should be made in order to achieve the desired results, in a move towards a future paradigm of a DNN-based artificial intelligence (AI) assisted chemistry research. This naturally leads to the second drawback of deep learning as the inevitable question surfaces - How do we know that the deep learning model is learning the correct physics or chemistry?

We will concede that in its current implementation deep learning algorithms is still a black box and interrogating what it "learns" is an extremely challenging task. Nevertheless, black box algorithms such as SVM and RF are also used in several computational chemistry applications, notably in examples where they are used primarily as a tool, and/or for prediction of properties that are so complex that even a first principles understanding of the problem will not necessarily aid in its prediction. We acknowledge that in order to advance deep learning to be more than just another tool in the chemist's toolkit, and for it to gain more widespread applicability and adoption for scientific research, it is evident that improvement in interpretability of DNN is of paramount interest. While interpretability of neural networks has historically not been a strong research focus for practitioners in this field, it is noteworthy that several recent developments on improving interpretability has been reported.[148,149] Other viable options include the use of different neural network based machine learning models, such as influence-relevance voters (IVR) that are designed for interpretability. As demonstrated on a few computational chemistry applications from the work of Baldi and co-workers[150,151], the IRV is a low-parameter neural network which refines a k-nearest neighbor classifier by nonlinearly combining the influences of a chemical's neighbors in the training set. IRV influences are decomposed, also nonlinearly, into a relevance component and a vote component. Therefore, the predictions of the IRV is by nature transparent, as the exact data used to make a prediction can be extracted from the network by examining each prediction's influences, making it closer to a "white-box" neural network method.[150,151]

## 8. Conclusion

Unlike traditional machine learning algorithms currently used in computational chemistry, deep learning distinguishes itself in its use of a hierarchical cascade of non-linear functions. This allows it to learn representations and extract out the necessary features from raw unprocessed data needed to predict the desired physicochemical property of interest. It is this distinguishing feature that has enabled deep learning to make significant and transformative impact in its "parent" field of speech recognition and computer vision. In computational chemistry, its impact is more recent and more preliminary. Nevertheless, based on the results from a number of recent studies, we have noted the broad application of deep learning in many sub-fields of computational chemistry, including computer aided drug design, computational structural biology, quantum chemistry and materials design. In almost all applications we have examined, the performance of DNN-based model is frequently superior to traditional machine learning algorithms.

As the complexity of the problem increases that enables the application of multi-task learning (i.e. more predictions of different properties are required), and as the size of the dataset increases, we have also seen deep learning progressing from frequently outperforming to always outperforming traditional machine learning models. In addition, some preliminary findings indicate that explicitly engineered features such as molecular descriptors may not be necessary to construct a high performing DNN model, and simpler representations in the form of molecular fingerprint or coulomb matrices may suffice. This is because of DNN's ability to extract out its own features through its hidden layers. There is even indication that the features "learned" by DNNs correspond to actual chemical concepts such as toxicophores. Coupled with recent research on improving interpretability of neural networks, it suggest that the future role of DNN in computational chemistry may not just be only a high-performance prediction tool, but perhaps as a hypothesis generation device as well.

**Acknowledgement**

The authors thank Dr. Nathan Baker for critically reviewing the manuscript and providing helpful comments and insights on machine learning for computational chemistry applications. This research was funded by the Pacific Northwest Laboratory Directed Research and Development (LDRD) Program and the Linus Pauling Distinguished Postdoctoral Fellowship.